\documentclass[letterpaper]{article} 
\usepackage[preprint]{aaai2027}  
\usepackage[hyphens]{url}
\usepackage{graphicx}
\usepackage{natbib}
\usepackage{caption}
\usepackage{amsmath,amssymb}
\usepackage{booktabs}
\usepackage{multirow}
\usepackage{subcaption}
\usepackage{xcolor}
\usepackage{tikz}
\usepackage{algorithm}
\usepackage{algpseudocode}
\usetikzlibrary{arrows.meta,positioning,fit,backgrounds}
\definecolor{sotaC}{RGB}{11,107,45}   
\definecolor{sndC}{RGB}{224,138,52}   
\definecolor{refC}{RGB}{130,130,130}  
\newcommand{\sota}[1]{\textbf{\textcolor{sotaC}{#1}}}
\newcommand{\snd}[1]{\textcolor{sndC}{#1}}
\newcommand{\refc}[1]{\textcolor{refC}{#1}}
\providecommand{\ConeGaussian}{ConeGaussian}
\newlength{\metricslot}
\newcommand{\nometric}{%
  \settowidth{\metricslot}{.000}%
  \makebox[\metricslot][c]{--}%
}
\newcommand{\titleicon}{%
  \raisebox{-0.28\height}{%
    \includegraphics[
      height=1.05cm
    ]{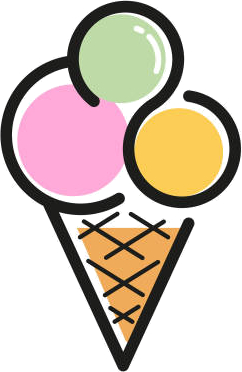}%
  }%
}

\DeclareRobustCommand{\ConeGaussian}{%
  \mbox{\textsc{\bfseries ConeGaussian}}%
}

\nocopyright

\title{\titleicon \hspace{0.2em} \ConeGaussian: Anti-Aliased Gaussian Ray-Tracing for Generic Central Cameras}
\author{%
\resizebox{\textwidth}{!}{%
Deheng Zhang\textsuperscript{1,*,$\dagger$}, Letian Shi\textsuperscript{1,*}, Runyi Yang\textsuperscript{1},
Zhendong Li\textsuperscript{1}, Lei Sun\textsuperscript{1}, Kanzhi Wu\textsuperscript{2}, Ajad Chhatkuli\textsuperscript{1,$\ddagger$}, Danda Pani Paudel\textsuperscript{1}, Luc Van Gool\textsuperscript{1,$\ddagger$}%
}
}
\affiliations{
\textsuperscript{1}INSAIT, Sofia University ``St. Kliment Ohridski''\\
\textsuperscript{2}vivo Mobile Communication Co., Ltd., Shenzhen, China\\
\textsuperscript{*}Equal contribution (co-first authors).\quad
\textsuperscript{$\dagger$}Project lead.\quad
\textsuperscript{$\ddagger$}Equal supervision.
}

\begin{document}
\maketitle

\begin{abstract}
In rendering, a camera is a sampling operator that maps each finite pixel to a bundle of rays. Different camera models change the geometry of this bundle, thus making a unified and faithful rendering formulation challenging. Consequently, gaussian ray tracing supports generic cameras (with optical center) through their inverse ray mappings, yet typically reduces every pixel to a single center ray. This ignores the camera-dependent pixel footprint, causing aliasing under minification, while unconstrained Gaussians expose unsupported frequencies under magnification. We present ConeGaussian, a camera-model-agnostic anti-aliasing framework for Gaussian ray-based rendering. Instead of defining the pixel filter on a camera-specific image plane, ConeGaussian constructs an anisotropic footprint directly from neighboring rays produced by the camera’s native inverse mapping. 
We derive a closed-form response under a locally linear, depth-local, moment-matched approximation of the finite pixel footprint, while the same geometry defines a per-Gaussian training-frequency floor. Notably, by construction, our filtering principle can be used unmodified across calibrated central camera models and multiple Gaussian ray-rendering backbones. Additionally, unlike in mip-splatting, our scene-space frequency floor and filtering enable trivial composition at render time allowing us to remove excess blurring. On pinhole and strongly distorted fisheye captures, ConeGaussian consistently improves two distinct ray-based backbones, by up to 4.3 dB at 
1/8 resolution, and reduces fisheye LPIPS by 30\% where a perspective screen-plane footprint formulations are not directly applicable. Code and checkpoints will be publicly available.

\vspace{-1em}
\end{abstract}

\section{Introduction}

An effective 3D scene representation should support the rendering of the same scene across different camera types in good quality. Perspective, fisheye and omnidirectional cameras offer different trade-offs in angular resolution, field of view, and blind-spot coverage \citep{liao2024fisheyegs,li2024omnigs}. 3D Gaussian Splatting (3DGS)~\citep{kerbl20233dgs} is attractive for this purpose, providing photorealistic rendering at interactive rates and supporting large-scale reconstruction and robotic navigation \citep{yan2024streetgaussians,gaussnav,splatnav}. However, its standard renderer projects each anisotropic Gaussian onto a perspective image plane using a local affine EWA approximation~\citep{zwicker2001ewa}, coupling the renderer to a specific projection model and degrading under wide-angle or strongly distorted cameras. More fundamentally, each pixel integrates a finite, camera-dependent bundle of rays rather than only its center ray, and this footprint can become elongated and sheared toward the image periphery. Generic-camera rendering and anti-aliasing must therefore be addressed together using the camera's native rays and their finite pixel footprints.

\begin{figure}[t]
  \centering
  \includegraphics[
    width=0.48\textwidth
  ]{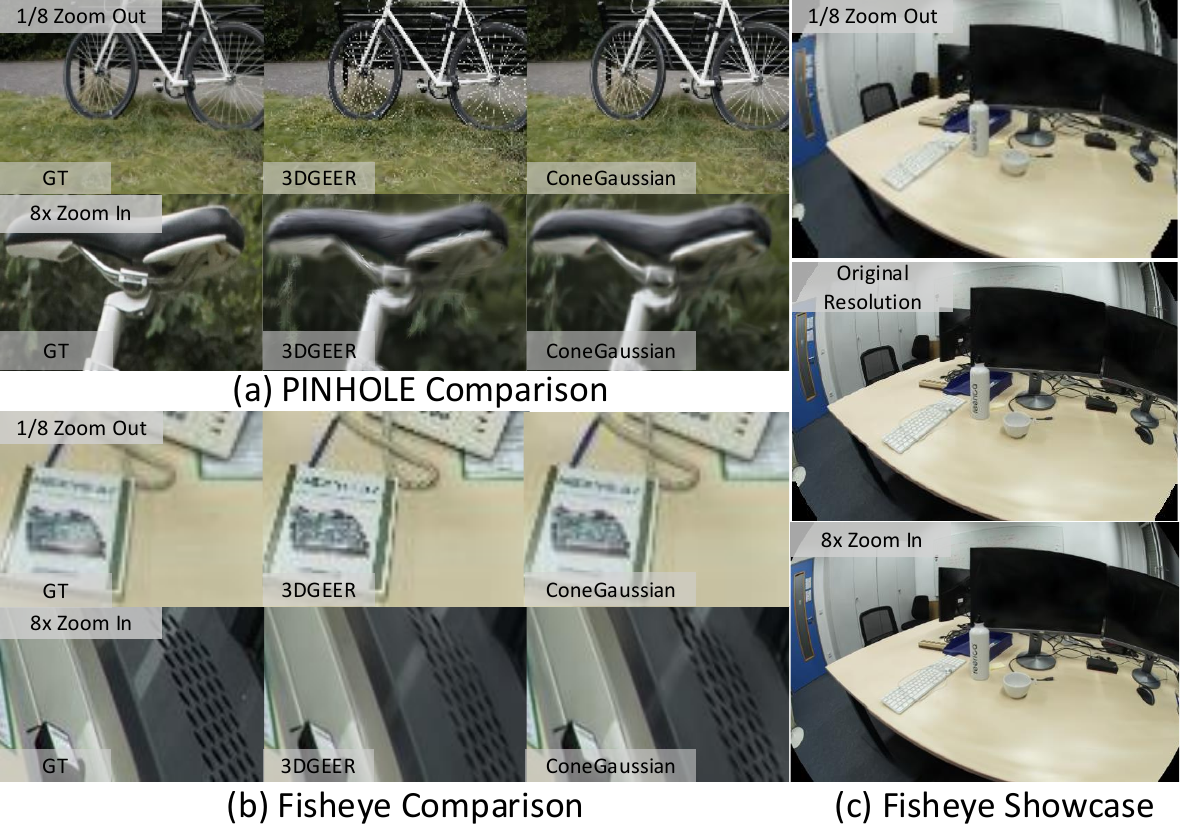}
  \vspace{-2em}

\caption{\ConeGaussian{} suppresses aliasing at $\tfrac18$ minification and preserves detail at $\times8$ magnification for both pinhole and fisheye cameras.}
  
  \vspace{-2em}
  \label{fig:teaser}
\end{figure}

Recent Gaussian renderers bypass this screen-space dependence by evaluating primitives directly in 3D or along camera rays \citep{moenne20243dgrt,wu20253dgut,huang20253dgeer}. In particular, 3DGEER~\citep{huang20253dgeer} provides a closed-form ray response, allowing calibrated central cameras to share the same renderer and differ only in their inverse projection. However, it still treats each finite pixel as a single center-ray sample. Conversely, anti-aliasing methods such as Mip-Splatting and Analytic-Splatting \citep{yu2024mip,liang2024analytic} model the pixel footprint only after projection onto a perspective image plane. VKRayGS~\citep{bulo2025raygs} introduces filtering into ray-based rendering, but approximates the footprint isotropically, discarding its shear and sizes with the paraxial $z/f$, which over-dilates an off-axis pixel. Thus, existing methods support either generic-camera ray evaluation or finite-pixel filtering, but not an anisotropic footprint derived directly from the camera's native rays.

We introduce \ConeGaussian, an anti-aliased Gaussian ray-tracing framework that derives each pixel filter directly from the camera's native rays. For every pixel--Gaussian pair, neighboring-ray differentials are transported to the Gaussian's maximum-response depth and represented as an anisotropic rank-two covariance in the Gaussian-whitened, ray-orthogonal plane. A moment-matched Gaussian then yields a closed-form filtered response. Since the construction requires only an inverse ray mapping, the same formulation applies to any calibrated central camera. To additionally suppress unsupported frequencies under magnification, we derive a per-Gaussian training-frequency floor from the same footprint geometry and add only the render-time variance not already covered by this floor, jointly handling minification and magnification without redundant smoothing.

We validate \ConeGaussian{} against densely supersampled pixel responses and on pinhole and strongly distorted fisheye benchmarks. As shown in Fig.~\ref{fig:teaser}, \ConeGaussian{} reduces aliasing under $1/8\times$ downsampling while preserving fine structures under $8\times$ magnification for both pinhole and fisheye cameras, including near the highly distorted fisheye periphery. The zoom in/out settings are further introduced in Sec.~\ref{sec:exp}. Its anisotropic footprint more accurately approximates finite-pixel integration than center-ray or isotropic filtering, especially at large eccentricities. Across multiple settings, a single implementation consistently improves the ray-based baseline and supports cross-camera-model rendering without changing the rendering formulation. Because the footprint is derived directly from the rendering rays rather than a backbone-specific screen-space projection, it can be incorporated into compatible Gaussian ray-based renderers. 
We instantiate it on both 3DGEER~\citep{huang20253dgeer} and 3DGUT~\citep{wu20253dgut}, demonstrating its portability across two distinct Gaussian ray-based response formulations. 
Our contributions are as follows:
\begin{itemize}


    \item \emph{Ray-native anisotropic filtering across cameras and Gaussian ray renderers.}
    We derive a ray-native anisotropic pixel footprint that enables a unified filtering formulation across calibrated central camera models and multiple Gaussian ray-based renderers. We provide the theoretical derivations and error bound analysis for the footprint filter in the Appendix.
    
    \item \emph{A unified treatment of sampling and representation band-limits.}
    The same ray-footprint geometry defines a per-Gaussian training-frequency floor, while the marginal composition rule composes it with the render-time footprint without redundant smoothing.

    \item \emph{A comprehensive evaluation across cameras, scales, and backbones.}
    We validate performance improvements in multi-scale and cross-camera rendering across multiple Gaussian ray-based renderers.
\end{itemize}

\begin{figure*}[t]
\centering
\includegraphics[width=\textwidth]{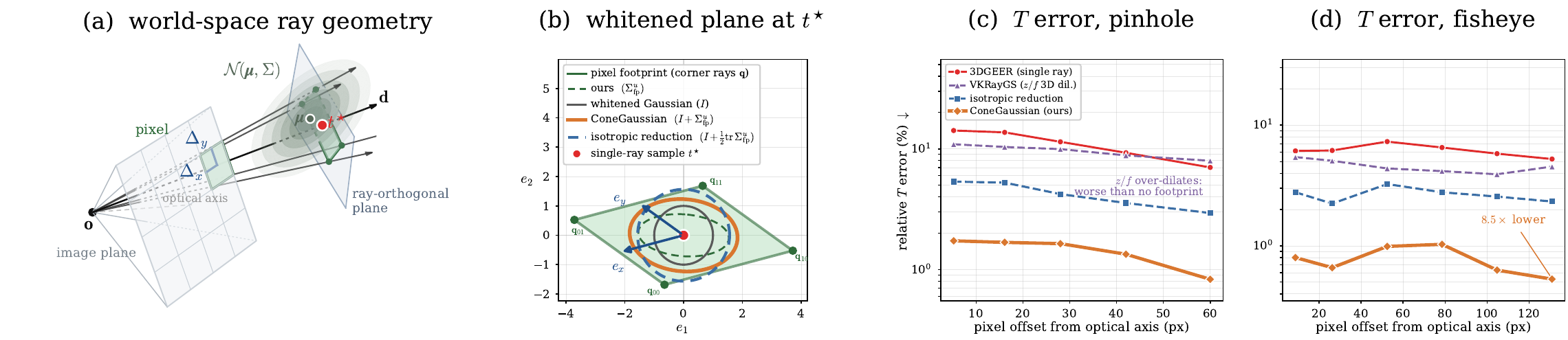}
\caption{\textbf{The \ConeGaussian{} pixel-footprint filter.}
(a) Pixel-edge ray differentials form a world-space footprint at the
maximum-response depth $t^\ast$.
(b) Whitening yields an anisotropic rank-two covariance; \ConeGaussian{}
preserves its orientation and aspect ratio, unlike single-ray
3DGEER~\citep{huang20253dgeer} and isotropic
VKRayGS-style filtering~\citep{bulo2025raygs}.
(c,d) Against a $32{\times}32$ supersampled reference, the anisotropic
filter achieves the lowest response error for both pinhole and fisheye
cameras, especially off-axis.
Beyond this single-Gaussian response test, we also validate multi-Gaussian
occlusion compositing against dense sub-ray references; see the Appendix for
details.}
\label{fig:geometry}
\end{figure*}

\section{Related Work}

\paragraph{3DGS Ray-Tracing and Anti-Aliasing.}
Pixel-footprint filtering is a classic anti-aliasing principle: texture mapping integrates over an elliptical footprint derived from neighboring ray differentials~\citep{heckbert1989fundamentals,igehy1999tracing}. We extend this construction to ray--Gaussian rendering by transporting exact edge-ray differentials to each Gaussian and convolving its response with the resulting anisotropic footprint. EWA splatting~\citep{zwicker2001ewa} introduced footprint filtering for point rendering, whereas 3DGS omits its screen-space band-limiting filter for efficiency. Related methods include conical-frustum integration in Mip NeRF~\citep{barron2021mipnerf,barron2022mipnerf360,barron2023zipnerf}, scale-specific primitives in Multi-Scale 3DGS~\citep{yan2024multiscale}, 3D frequency constraints and 2D mip filtering in Mip-Splatting~\citep{yu2024mip}, and analytic pixel integration in Analytic-Splatting~\citep{liang2024analytic}. Closest to ours, VKRayGS~\citep{bulo2025raygs} approximates the pixel footprint using isotropic world-space dilation based on the paraxial $z/f$ interval, sacrificing anisotropy for hardware efficiency while remaining tied to perspective rasterization. In contrast, our anisotropic footprint supports a broad range of calibrated central camera models.

\paragraph{Exact / Generic-Camera Gaussian Rendering.}
Screen-space methods inherit 3DGS's local affine projection, which degrades under wide fields of view and strong distortion. Ray-based methods avoid this approximation and differ mainly in how they define the ray--Gaussian response. Maximum-response approaches evaluate the Gaussian at the point of minimum Mahalanobis distance: RayGauss~\citep{blanc2025raygauss} and 3DGRT~\citep{moenne20243dgrt} uses BVH-based ray tracing, 3DGUT~\citep{wu20253dgut} applies an unscented transform, and Gaussian Opacity Fields~\citep{yu2024gof} evaluate opacity at ray--Gaussian intersections. Integral-based methods instead compute the response along the ray in closed form: VKRayGS~\citep{bulo2025raygs} marginalizes Gaussians onto the ray-orthogonal plane, while 3DGEER~\citep{huang20253dgeer} derives the exact ray integral from the minimum Mahalanobis distance and uses Particle Bounding Frustums to support generic cameras without BVH intersections. EVER~\citep{mai2024ever} exactly renders constant-density ellipsoids, whereas FisheyeGS~\citep{liao2024fisheyegs} adapts splatting to fisheye projection. None models the anisotropic pixel footprint, thus, we introduce it while preserving its exactness and camera generality.  

\section{Preliminaries: Ray-Tracing 3DGS}
\label{sec:prelim}

Previous Gaussian ray-based rendering methods, such as 3DGUT~\citep{wu20253dgut} and 3DGEER~\citep{huang20253dgeer}, evaluate Gaussian primitives directly along camera rays, with 3DGEER further deriving a closed-form solution for the Gaussian ray integral. Rays are cast from the camera's optical center
$\mathbf{o}\in\mathbb{R}^3$: any pixel $(x,y)$ of a camera with inverse projection $\Pi^{-1}$ yields a world ray $(\mathbf{o},\mathbf{d})$
with direction unit vector $\mathbf{d}$, written in world coordinates as
$\mathbf{r}(t)=\mathbf{o}+t\mathbf{d}$. A Gaussian $\mathcal{G}_k$ has mean
$\boldsymbol\mu\in\mathbb{R}^3$, covariance
$\boldsymbol\Sigma=\mathbf{R}\,\mathbf{S}^2\mathbf{R}^\top$ (rotation
$\mathbf{R}$, scales $\mathbf{S}=\mathrm{diag}(s_1,s_2,s_3)$) and opacity
$\sigma$. A world-space point $\mathbf{x}$ is mapped to the whitened Gaussian
coordinate $\mathbf{u}$ by
$\mathbf{u}=\mathbf{W}(\mathbf{x}-\boldsymbol\mu)$ with
$\mathbf{W}=\mathbf{S}^{-1}\mathbf{R}^\top$, which maps the Gaussian to the
unit isotropic ball. Whitening the ray with direction
$\mathbf{d}_u=\mathbf{W}\mathbf{d}$ and center $\mathbf{o}_u=\mathbf{W}(\mathbf{o}-\boldsymbol\mu)$ gives
\begin{equation}
\mathbf{r}_u(t)=\mathbf{o}_u+t\,\mathbf{d}_u .
\end{equation}

Because whitening maps $\boldsymbol\Sigma$ to the identity, the Gaussian
response along the transformed ray $\mathbf r_u(t)=\mathbf o_u+t\mathbf d_u$
is a one-dimensional Gaussian in the ray parameter $t$. Its maximum occurs at
$
t^\ast=
-\frac{\langle \mathbf o_u,\mathbf d_u\rangle}{\|\mathbf d_u\|^2}
$
, and the corresponding minimum squared
Mahalanobis distance, \textup{i.e.}, the squared perpendicular distance from the ray to
the Gaussian centre, is
\begin{equation}
D^2
=
\left\|
\mathbf o_u+t^\ast\mathbf d_u
\right\|^2
=
\frac{\|\mathbf d_u\times\mathbf o_u\|^2}{\|\mathbf d_u\|^2}.
\label{eq:geer_normalized}
\end{equation}

We therefore define a unified weighted ray response
\begin{equation}
T[w]
=
\sigma\!\int_{\mathbb R}
w(t)\,
\mathcal G_{\boldsymbol\mu,\boldsymbol\Sigma}
\bigl(\mathbf o+t\mathbf d\bigr)\,dt
=
C[w]\,\sigma\,e^{-\frac12D^2},
\label{eq:weighted_ray_response}
\end{equation}
where
$\mathcal G_{\boldsymbol\mu,\boldsymbol\Sigma}(\mathbf x)
:=
\frac{1}{Z_G\sqrt{\det\boldsymbol\Sigma}}\,
\exp\!\left(
-\frac12
(\mathbf x-\boldsymbol\mu)^\top
\boldsymbol\Sigma^{-1}
(\mathbf x-\boldsymbol\mu)
\right)
$
is a Gaussian with mean $\boldsymbol\mu$, covariance $\boldsymbol\Sigma$, and
constant normalizer $Z_G$, and
$C[w]$ is a longitudinal aggregation factor.

Two representative instances of this formulation are 3DGUT and 3DGEER,
corresponding to $w_{\mathrm{GUT}}(t)=\delta(t-t^\ast)$ and
$w_{\mathrm{GEER}}(t)=1$, respectively.


Following vanilla 3DGS, Gaussians are sorted by their max-response depths $t_k^\ast$ and composited front-to-back:
\begin{equation}
\mathbf{c}(x,y)=\sum_k \mathbf{c}_k T_k \prod_{j<k}(1-T_j),
\label{eq:blend}
\end{equation}
where $T_k$ is the ray--Gaussian response and $\mathbf{c}_k$ its view-dependent colour.


\section{Methodology: \ConeGaussian}
\label{sec:method}

Although previous Gaussian ray-tracing methods integrate each Gaussian along a viewing ray, they
represent each pixel by a single center ray. Under zoom-out, where scene
details are projected below the pixel sampling resolution, frequencies
beyond the Nyquist limit are not filtered and therefore alias into the
sampled image. Conversely, under zoom-in, the absence of a 3D frequency
constraint allows Gaussians to encode frequencies beyond those supported by
the training views, producing high-frequency artifacts when rendered at a
higher sampling rate. We address scale-dependent aliasing from two
complementary directions: (1) for zoom-out, we model each pixel as a finite
viewing cone rather than an infinitesimal ray and filter the Gaussian
response over its local cross-section; (2) for zoom-in, we constrain the
minimum 3D Gaussian scale to band-limit the scene representation and
suppress unsupported high-frequency components.

\subsection{Pixel-Footprint Filtering}
\label{sec:prefilter}

Our method is based on three assumptions. We further detailed and analyzed the assumptions in the Appendix. 
\begin{itemize}
\itemsep2pt
\item \textbf{Assumption 1 (Local ray-linearity).} The ray-direction field is
locally linear within a pixel, so its footprint is approximated by a
parallelogram.
\item \textbf{Assumption 2 (Depth locality).} The ray-orthogonal footprint
changes little over the effective ray-integration range of a Gaussian, so it
can be evaluated at the maximum-response depth $t^\ast$.
\item \textbf{Assumption 3 (Gaussian moment matching).} The uniform footprint
distribution is approximated by a moment-matched Gaussian that preserves its
mean and covariance.
\end{itemize}

\paragraph{Pixel ray bundle.}
A pixel represents a finite bundle of rays rather than a single center ray.
Let $\mathcal P=[-\frac12,\frac12]^2$ denote the pixel domain and
$\mathbf d(\boldsymbol\xi)=\Pi^{-1}(\mathbf p+\boldsymbol\xi)$ the ray
direction of the offset $\boldsymbol\xi\in\mathcal P$ from the pixel centre
$\mathbf p=(x,y)$. The pixel response is the average of the Gaussian response
over this ray bundle:
\begin{equation}
T
=
\frac{\sigma}{|\mathcal P|}
\int_{\mathcal P}\int_{\mathbb R}
w(t)\mathcal G_{\boldsymbol\mu,\boldsymbol\Sigma}
\bigl(\mathbf o+t\mathbf d(\boldsymbol\xi)\bigr)
\,dt\,d\boldsymbol\xi .
\label{eq:bundle}
\end{equation}

\paragraph{Footprint covariance and filtered response.}
To estimate Eq.~\ref{eq:bundle}, we define the ray-orthogonal plane and the parallelogram projection (detail proof in the Appendix). With
$\mathbf M_{\!\perp}=\mathbf I-\mathbf d_u\mathbf d_u^\top/
(\mathbf d_u^\top\mathbf d_u)$ the orthogonal projector onto the plane
perpendicular to the whitened ray (ray-orthogonal plane), the differentials are in practice taken from the camera model rather
than an analytic Jacobian: with the half-pixel offsets
$\boldsymbol\Delta_x=(\tfrac12,0)$, $\boldsymbol\Delta_y=(0,\tfrac12)$ (different estimation analysis is in Appendix) and the
edge rays $\mathbf{d}_i=\Pi^{-1}(\mathbf{p}+\boldsymbol\Delta_i), \qquad i\in\{x,y\}$ from the
\emph{same} camera model as $\mathbf{d}$, the whitened and projected half-edges are vectors: 
\begin{equation}
\mathbf{e}_i
=
t^\ast\,\mathbf{M}_{\!\perp}\mathbf{W}
\big(\mathbf{d}_i-\mathbf{d}\big),
\qquad i\in\{x,y\}.
\end{equation}
The half-edge vectors $\mathbf{e}_x$ and $\mathbf{e}_y$ span the parallelogram
(Fig.~\ref{fig:geometry}(a))
\begin{equation}
\Omega_{\mathrm{fp}}^{u}
=
\left\{
a\,\mathbf{e}_x+b\,\mathbf{e}_y
\;\middle|\;
a,b\in[-1,1]
\right\} .
\label{eq:footprint_set}
\end{equation}
A point uniformly distributed over $\Omega_{\mathrm{fp}}^{u}$ has zero mean.
Since a scalar uniformly distributed over $[-1,1]$ has variance $1/3$, each
half-edge vector $\mathbf{e}$ contributes the covariance
$\mathbf{e}\mathbf{e}^\top/3$. Thus, we define the footprint covariance in the whitened space $\boldsymbol\Sigma_{\mathrm{fp}}^{u}$
\begin{equation}
\boldsymbol\Sigma_{\mathrm{fp}}^{u}=\tfrac{1}{3}\big(\mathbf{e}_x\mathbf{e}_x^\top
+\mathbf{e}_y\mathbf{e}_y^\top\big),
\qquad
\boldsymbol\Sigma_{\mathrm{fp}}^{u}\,\mathbf{d}_u=\mathbf{0} .
\label{eq:sigmafp}
\end{equation}
The inner average over $\Omega_{\mathrm{fp}}^{u}$ is therefore a convolution with
the normalized uniform footprint kernel
$K_{\mathrm{fp}}(\mathbf z_u)=\mathbf 1_{\Omega_{\mathrm{fp}}^{u}}(\mathbf z_u)/
|\Omega_{\mathrm{fp}}^{u}|$, \emph{i.e.,}\
$\widetilde{\mathcal G}_{\mathbf0,\mathbf I}
=\mathcal G_{\mathbf0,\mathbf I}*K_{\mathrm{fp}}$.
Following Assumption~3 we approximate $K_{\mathrm{fp}}$ by the Gaussian
$\mathcal{G}_{\mathbf{0},\boldsymbol\Sigma_{\mathrm{fp}}^{u}}$ of the same
mean and covariance, so that the convolution becomes a covariance sum,
$\widetilde{\mathcal G}_{\mathbf0,\mathbf I}
\approx\mathcal G_{\mathbf0,\mathbf I}*
\mathcal G_{\mathbf{0},\boldsymbol\Sigma_{\mathrm{fp}}^{u}}$.
With
$\mathbf{q}=\mathbf{M}_{\!\perp}\mathbf{o}_u$ the vector from the Gaussian
centre to the closest point of the ray, \emph{i.e.,}\ the perpendicular ray-to-centre
offset already appearing in Eq.~\eqref{eq:geer_normalized} through
$\lVert\mathbf{q}\rVert^2=D^2$, we obtain the final Gaussian response $T$ and
the filtered covariance $\mathbf{A}$ of the whitened Gaussian,
\begin{equation}
T
=\frac{\sigma}{\sqrt{\det\mathbf{A}}}\;
e^{-\frac12\,\mathbf{q}^\top\mathbf{A}^{-1}\mathbf{q}} ,
\qquad
\mathbf{A}=\mathbf{I}+\boldsymbol\Sigma_{\mathrm{fp}}^{u} .
\label{eq:ours}
\end{equation}
Since $\boldsymbol\Sigma_{\mathrm{fp}}^{u}$ lies entirely in the ray-orthogonal
plane, $\mathbf A$ remains unchanged along $\mathbf d_u$, allowing the
filtering terms to be evaluated without explicitly constructing a plane basis
to convert the 3D representation into 2D ray-orthogonal coordinates. Our
formulation therefore extends the 2D mip-filtering principle of Mip-Splatting
to calibrated central camera models by expressing the pixel footprint in a
ray-orthogonal frame.

The isotropic reduction of Eq.~\eqref{eq:ours}, which applies where
footprints stay near-isotropic, and a discussion of what is
linearized in our construction and in 3DGS are deferred to the Appendix.

\subsection{Regularizing Representation Frequencies}
\label{sec:marginal}

Eq.~\eqref{eq:ours} addresses rendering-time aliasing by integrating over the
pixel footprint. Independently, the Gaussian representation may contain
frequencies beyond the training resolution, which can appear as spurious fine
structures under magnification. We therefore impose a filter floor on each
Gaussian, attenuating high-frequency content beyond the finest resolution
observed.

\paragraph{Footprint floor filter.} Following Mip-Splatting, we associate each Gaussian with a scalar $\rho^2$, and apply an isotropic scene-space
dilation,
\begin{equation}
\boldsymbol\Sigma_{\mathrm{eff}}
=
\boldsymbol\Sigma + \rho^2\mathbf I,
\qquad
\sigma_{\mathrm{eff}}
=
\sigma
\prod_{i=1}^{3}
\frac{s_i}{\sqrt{s_i^2+\rho^2}},
\label{eq:floor}
\end{equation}
at both training and test time. Here,
$\boldsymbol\Sigma_{\mathrm{eff}}$ denotes the effective scene-space Gaussian
covariance used for training and rendering after applying the training-scale
representation regularization. The covariance dilation, defined as footprint floor filter, is equivalently
obtained by changing each principal-axis scale from $s_i$ to
$\sqrt{s_i^2+\rho^2}$, while the product term compensates for the corresponding
change in the $3$D Gaussian volume.

It is evaluated from the finest footprint with which each Gaussian is observed
during training, using the same ray-based footprint covariance that drives the
render-time filter,
\begin{equation}
\rho^2
=
\min_{v\in\mathcal V}
\tfrac12\operatorname{tr}\boldsymbol\Sigma_{\mathrm{fp}}^{(v)} ,
\label{eq:rho2}
\end{equation}
where $\boldsymbol\Sigma_{\mathrm{fp}}^{(v)}$ denotes the scene-space
footprint covariance associated with the pixel onto which the Gaussian
center projects in training view $v$, obtained using the same neighbour-ray
construction as Eq.~(9) but before whitening, and
$\mathcal V$ is the set of views in which the Gaussian is visible; the
half-trace is the isotropic (per-axis) reduction of that covariance. 
The minimum over visible training views therefore sets the
smallest representation scale supported by the training observations. Replacing $\boldsymbol\Sigma_k$ with $\boldsymbol\Sigma_{k,\mathrm{eff}}$ changes the whitening transform and the corresponding ray-orthogonal geometry, and  $\boldsymbol\Sigma_{\mathrm{fp}}^{u}$ is recomputed accordingly as $\boldsymbol\Sigma_{\mathrm{fp},\mathrm{eff}}^{u}$.
 

\paragraph{Marginal composition.}
The pixel-footprint filter and the footprint-floor filter are
complementary: the former adapts filtering to the sampling footprint of the
current view, while the latter regularizes the Gaussian representation at the
training scale. Directly adding the two, however, may introduce redundant
smoothing near the training resolution. We therefore retain only the footprint
variance beyond the baked floor,
\begin{equation}
\boldsymbol\Sigma_{\mathrm{fp,render}}^{u}
=
\mathbf W
\left[
\mathbf W^{-1}
\boldsymbol\Sigma_{\mathrm{fp},\mathrm{eff}}^{u}
\mathbf W^{-\top}
-
\rho^2\mathbf I
\right]_{+}
\mathbf W^\top ,
\label{eq:marginal}
\end{equation}
where $[\cdot]_+$ denotes a positive-semidefinite clamp. Since the world-space
footprint covariance
$\left(\mathbf W^{-1}\boldsymbol\Sigma_{\mathrm{fp},\mathrm{eff}}^{u}\mathbf W^{-\top}\right)$
is rank-2, the clamp acts direction-wise: footprint variance already covered
by the floor is suppressed, while only the excess variance is retained.
Consequently, no additional render-time filtering is introduced along
directions already covered by the floor, whereas larger footprints contribute
only the additional variance required by the current view. This avoids
redundant smoothing while preserving view-dependent anti-aliasing.

\section{Experiments}
\label{sec:exp}

\paragraph{Datasets.}
We evaluate on three real-scene benchmarks spanning indoor and outdoor scenes and two central camera models. For \emph{fisheye} rendering, we use ScanNet++~\citep{yeshwanth2023scannetpp} and Zip-NeRF~\citep{barron2023zipnerf}; for \emph{pinhole} rendering, we use Mip-NeRF 360~\citep{barron2022mipnerf360}. We additionally report results on the synthetic Blender benchmark in the Appendix, giving four benchmarks in total. For the real-scene datasets, every eighth image is held out for evaluation. We report PSNR, SSIM, and LPIPS. On fisheye images, all metrics are computed within a $3$-pixel-eroded valid-domain mask to exclude boundary artifacts. 

\begin{table*}[htbp]
\centering\small
\setlength{\tabcolsep}{3pt}
\begin{tabular}{ll ccccc}
\toprule
Dataset & Method & $1$ & $\tfrac12$ & $\tfrac14$ & $\tfrac18$ & Avg.\\
\midrule
\multirow{4}{*}{\shortstack[l]{\emph{ScanNet++}\\[1pt]\scriptsize fisheye}}
& 3DGUT
& \snd{29.33}\,/\,.915\,/\,.248 & \snd{30.38}\,/\,\snd{.934}\,/\,.193 & \snd{31.47}\,/\,\snd{.954}\,/\,.112 & \snd{30.24}\,/\,.953\,/\,.078 & \snd{30.36}\,/\,.939\,/\,.158\\
& ConeGaussian {\scriptsize\textit{w.}\,3DGUT}
& \sota{29.67}\,/\,\sota{.921}\,/\,\sota{.239} & \sota{30.61}\,/\,\sota{.938}\,/\,\sota{.185} & \sota{31.66}\,/\,\sota{.956}\,/\,\sota{.108} & \sota{31.65}\,/\,\sota{.963}\,/\,\snd{.059} & \sota{30.90}\,/\,\sota{.944}\,/\,\sota{.148}\\
& 3DGEER
& 27.59\,/\,.911\,/\,.257 & 28.53\,/\,.929\,/\,.202 & 28.53\,/\,.946\,/\,.123 & 28.02\,/\,.946\,/\,.084 & 28.17\,/\,.933\,/\,.167\\
& ConeGaussian {\scriptsize\textit{w.}\,3DGEER}
& 27.65\,/\,\snd{.917}\,/\,\snd{.243} & 28.29\,/\,.931\,/\,\snd{.189} & 29.27\,/\,.950\,/\,\snd{.111} & 29.93\,/\,\snd{.962}\,/\,\sota{.052} & 28.78\,/\,\snd{.940}\,/\,\snd{.149}\\
\cmidrule(lr){1-7}
\multirow{4}{*}{\shortstack[l]{\emph{Zip-NeRF}\\[1pt]\scriptsize fisheye}}
& 3DGUT
& 23.40\,/\,.781\,/\,.413 & 24.18\,/\,.835\,/\,.299 & 24.73\,/\,.867\,/\,.192 & 24.58\,/\,.865\,/\,.147 & 24.22\,/\,.837\,/\,.263\\
& ConeGaussian {\scriptsize\textit{w.}\,3DGUT}
& 23.48\,/\,.791\,/\,.403 & 24.22\,/\,.840\,/\,.292 & 24.82\,/\,.871\,/\,.185 & 25.14\,/\,.880\,/\,.128 & 24.42\,/\,.845\,/\,.252\\
& 3DGEER
& \snd{24.32}\,/\,\snd{.806}\,/\,\snd{.388} & \snd{25.02}\,/\,\snd{.854}\,/\,\snd{.268} & \snd{25.64}\,/\,\snd{.893}\,/\,\snd{.162} & \snd{25.45}\,/\,\snd{.895}\,/\,\snd{.121} & \snd{25.11}\,/\,\snd{.862}\,/\,\snd{.235}\\
& ConeGaussian {\scriptsize\textit{w.}\,3DGEER}
& \sota{24.42}\,/\,\sota{.815}\,/\,\sota{.377} & \sota{25.16}\,/\,\sota{.860}\,/\,\sota{.261} & \sota{25.89}\,/\,\sota{.898}\,/\,\sota{.153} & \sota{26.37}\,/\,\sota{.915}\,/\,\sota{.094} & \sota{25.46}\,/\,\sota{.872}\,/\,\sota{.221}\\
\cmidrule(lr){1-7}
\multirow{7}{*}{\shortstack[l]{\emph{Mip-NeRF 360}\\[1pt]\scriptsize Pinhole}}
& \refc{3DGS}
& \refc{26.55\,/\,.779\,/\,.274} & \refc{28.00\,/\,.854\,/\,.162} & \refc{28.51\,/\,.891\,/\,.102} & \refc{27.45\,/\,.888\,/\,.087} & \refc{27.63\,/\,.853\,/\,.156}\\
& \refc{Mip-Splatting$^\dagger$}
& \refc{27.20\,/\,.802\,/\,.244} & \refc{28.74\,/\,.870\,/\,.146} & \refc{29.90\,/\,.915\,/\,.090} & \refc{30.66\,/\,.944\,/\,.056} & \refc{29.12\,/\,.883\,/\,.134}\\
& \refc{Analytic-Splatting$^\dagger$}
& \refc{27.50\,/\,.808\,/\,.231} & \refc{28.99\,/\,.874\,/\,.132} & \refc{30.35\,/\,.919\,/\,.077} & \refc{31.21\,/\,.945\,/\,.051} & \refc{29.51\,/\,.887\,/\,.123}\\
\cmidrule(lr){2-7}
& 3DGUT
& 26.68\,/\,.768\,/\,.341 & 26.80\,/\,.788\,/\,.241 & 26.97\,/\,.823\,/\,.159 & 24.02\,/\,.716\,/\,.238 & 26.12\,/\,.774\,/\,.245\\
& ConeGaussian {\scriptsize\textit{w.}\,3DGUT}
& \snd{26.98}\,/\,\snd{.782}\,/\,.332 & 26.94\,/\,.798\,/\,.234 & 27.21\,/\,.831\,/\,.152 & 25.00\,/\,.762\,/\,.210 & 26.53\,/\,.793\,/\,.232\\
& 3DGEER
& 26.01\,/\,.739\,/\,\snd{.326} & \snd{27.55}\,/\,\snd{.820}\,/\,\snd{.212} & \snd{28.56}\,/\,\snd{.880}\,/\,\snd{.128} & \snd{26.74}\,/\,\snd{.860}\,/\,\snd{.128} & \snd{27.21}\,/\,\snd{.825}\,/\,\snd{.199}\\
& ConeGaussian {\scriptsize\textit{w.}\,3DGEER}
& \sota{27.05}\,/\,\sota{.792}\,/\,\sota{.275} & \sota{28.44}\,/\,\sota{.841}\,/\,\sota{.176} & \sota{30.03}\,/\,\sota{.905}\,/\,\sota{.095} & \sota{31.08}\,/\,\sota{.945}\,/\,\sota{.052} & \sota{29.15}\,/\,\sota{.871}\,/\,\sota{.149}\\
\bottomrule
\end{tabular}
\caption{\textbf{MTMT} (PSNR$\uparrow$/SSIM$\uparrow$/LPIPS$\downarrow$).
Models are trained and evaluated at
$\{1,\tfrac12,\tfrac14,\tfrac18\}$.
\refc{Gray} rows are author-reported screen-space pinhole-specialized baselines that \emph{\textbf{do not} support \textbf{generic }cameras}~\citep{yu2024mip,liang2024analytic} (reported here only for references).
\sota{Green}/\snd{orange} denote the best/second-best generalizable
ray-based methods per column and metric.}
\label{tab:ngauss}
\end{table*}

\paragraph{Scale convention.}
Throughout the experiments, \emph{zoom-in} and \emph{zoom-out} refer only to changes in image sampling resolution, rather than optical zoom or camera motion. We keep the camera model, pose, and view fixed, while scaling the camera calibration consistently with the image resolution. 

\paragraph{Zoom-in.}
We train a model only at the lowest resolution and render it at
$\times1$, $\times2$, and $\times4$, where $\times1$ denotes the training resolution. Increasing the rendering resolution reduces the angular footprint of each pixel and evaluates the model at sampling rates finer than those observed during training. This setting primarily evaluates the training-frequency floor used to suppress unsupported representation frequencies. For the fisheye benchmarks, we additionally report \emph{Corner $\times4$} to examine the strongly distorted image periphery. The $\times8$ setting is used only as an additional qualitative stress test.

\paragraph{Zoom-out.}
We train a model at native resolution and render it without retraining at linear resolution factors $\{1,\tfrac12,\tfrac14,\tfrac18\}$ relative to the native images. As the resolution decreases, each output pixel covers a larger bundle of rays. This setting evaluates rendering under minification and primarily tests the render-time pixel-footprint filter.

\paragraph{Multi-scale training and multi-scale testing (MTMT).}
Following the multi-scale protocol of Mip-NeRF~\citep{barron2021mipnerf}, a single model is trained using images at all four resolution levels $\{1,\tfrac12,\tfrac14,\tfrac18\}$ and evaluated separately at each level. Because every evaluation scale is represented during training, MTMT measures joint multi-scale fitting under mixed-resolution supervision rather than generalization to unseen resolutions.

\paragraph{Single-scale training and multi-scale testing (STMT).}
Following the single-scale-training protocol of Mip-Splatting~\citep{yu2024mip}, a model is trained at one resolution and evaluated at multiple resolutions without retraining. We instantiate STMT in two directions: native-to-coarser resolutions for zoom-out and lowest-to-finer resolutions for zoom-in. This protocol evaluates generalization to sampling rates that are not observed during training.

\subsection{Main Results}
\begin{table*}[t]
\centering\small
\setlength{\tabcolsep}{3pt}
\resizebox{\textwidth}{!}{
\begin{tabular}{ll ccccc}
\toprule
Dataset & Method & $1$ & $\tfrac12$ & $\tfrac14$ & $\tfrac18$ & Avg.\\
\midrule
\multirow{4}{*}{\shortstack[l]{\emph{ScanNet++}\\[1pt]\scriptsize fisheye}}
& 3DGEER (baseline)
& 27.89\,/\,\snd{.921}\,/\,.243 & \snd{28.18}\,/\,.930\,/\,.198 & 27.10\,/\,.930\,/\,.155 & 26.58\,/\,.918\,/\,.131 & 27.44\,/\,.925\,/\,.182\\
& VKRayGS\ $z/f$
& 26.20\,/\,.919\,/\,.244 & 25.89\,/\,.922\,/\,.205 & 21.57\,/\,.885\,/\,.193 & 20.54\,/\,.833\,/\,.185 & 23.55\,/\,.890\,/\,.207\\
& ConeGaussian (isotropic)
& \snd{27.96}\,/\,\snd{.921}\,/\,\snd{.242} & \sota{28.45}\,/\,\sota{.932}\,/\,\snd{.194} & \snd{27.34}\,/\,\snd{.937}\,/\,\snd{.137} & \snd{27.37}\,/\,\snd{.938}\,/\,\snd{.089} & \snd{27.78}\,/\,\snd{.932}\,/\,\snd{.165}\\
& ConeGaussian (anisotropic)
& \sota{27.97}\,/\,\sota{.922}\,/\,\sota{.239} & \sota{28.45}\,/\,\sota{.932}\,/\,\sota{.193} & \sota{28.25}\,/\,\sota{.942}\,/\,\sota{.128} & \sota{27.93}\,/\,\sota{.942}\,/\,\sota{.079} & \sota{28.15}\,/\,\sota{.934}\,/\,\sota{.160}\\
\cmidrule(lr){1-7}
\multirow{4}{*}{\shortstack[l]{\emph{Zip-NeRF}\\[1pt]\scriptsize fisheye}}
& 3DGEER (baseline)
& 24.13\,/\,.787\,/\,.409 & 24.92\,/\,.845\,/\,.278 & 25.87\,/\,.899\,/\,.150 & 25.82\,/\,.906\,/\,.104 & 25.19\,/\,.859\,/\,.235\\
& VKRayGS\ $z/f$
& 23.37\,/\,.786\,/\,.406 & 24.61\,/\,.846\,/\,.276 & \sota{26.00}\,/\,\sota{.902}\,/\,\sota{.148} & 25.92\,/\,.913\,/\,.092 & 24.97\,/\,.862\,/\,.230\\
& ConeGaussian (isotropic)
& \snd{24.16}\,/\,\snd{.794}\,/\,\snd{.401} & \snd{25.01}\,/\,\snd{.850}\,/\,\snd{.271} & \snd{25.97}\,/\,\snd{.901}\,/\,\sota{.148} & \snd{26.41}\,/\,\snd{.918}\,/\,\snd{.089} & \snd{25.39}\,/\,\snd{.866}\,/\,\snd{.227}\\
& ConeGaussian (anisotropic)
& \sota{24.25}\,/\,\sota{.795}\,/\,\sota{.398} & \sota{25.02}\,/\,\sota{.851}\,/\,\sota{.270} & 25.96\,/\,\snd{.901}\,/\,\snd{.148} & \sota{26.43}\,/\,\sota{.919}\,/\,\sota{.088} & \sota{25.41}\,/\,\sota{.867}\,/\,\sota{.226}\\
\cmidrule(lr){1-7}
\multirow{4}{*}{\shortstack[l]{\emph{Mip-NeRF 360}\\[1pt]\scriptsize pinhole}}
& 3DGEER (baseline)
& 27.03\,/\,.797\,/\,.300
& 27.68\,/\,\snd{.832}\,/\,.214
& 27.19\,/\,.836\,/\,.171
& 25.29\,/\,.784\,/\,.183
& 26.80\,/\,.812\,/\,.217\\
& VKRayGS\ $z/f$
& \snd{27.14}\,/\,.798\,/\,\snd{.298}
& 27.25\,/\,\snd{.832}\,/\,.219
& 25.68\,/\,.839\,/\,.175
& 22.96\,/\,.786\,/\,.183
& 25.76\,/\,.814\,/\,.219\\
& ConeGaussian (isotropic)
& \sota{27.17}\,/\,\snd{.799}\,/\,\sota{.296}
& \sota{28.10}\,/\,\sota{.843}\,/\,\snd{.207}
& \snd{28.80}\,/\,\sota{.884}\,/\,\snd{.135}
& \snd{28.53}\,/\,\snd{.903}\,/\,\snd{.096}
& \snd{28.15}\,/\,\sota{.857}\,/\,\snd{.183}\\
& ConeGaussian (anisotropic)
& 27.11\,/\,\sota{.803}\,/\,\sota{.296}
& \snd{28.06}\,/\,.829\,/\,\sota{.206}
& \sota{28.87}\,/\,\snd{.876}\,/\,\sota{.131}
& \sota{28.67}\,/\,\sota{.909}\,/\,\sota{.087}
& \sota{28.18}\,/\,\snd{.854}\,/\,\sota{.180}\\
\bottomrule
\end{tabular}
}
\caption{\textbf{Zoom-out}. Models are trained at native resolution and evaluated at $\{1,\tfrac12,\tfrac14,\tfrac18\}$ with the \textbf{STMT} setting~\citep{yu2024mip}.
}
\label{tab:zoomout}
\end{table*}

\begin{table*}[t]
\centering\small
\setlength{\tabcolsep}{3pt}
\resizebox{\textwidth}{!}{
\begin{tabular}{ll cccc@{\hspace{6pt}}|@{\hspace{6pt}}c}
\toprule
Dataset & Method & $\times1$ & $\times2$ & $\times4$ & Avg. & Corner\,$\times4$\\
\midrule
\multirow{4}{*}{\shortstack[l]{\emph{ScanNet++}\\[1pt]\scriptsize fisheye}}
& 3DGEER (baseline)
& 27.28\,/\,.944\,/\,.122 & 27.74\,/\,.918\,/\,.221 & 27.33\,/\,.895\,/\,.279 & 27.45\,/\,.919\,/\,.208 & 22.73\,/\,.871\,/\,\nometric\\
& 3DGEER (+ Mip-Splatting floor)
& \sota{29.49}\,/\,\sota{.951}\,/\,\snd{.111} & 27.19\,/\,.918\,/\,.216 & 27.25\,/\,.896\,/\,.272 & 27.98\,/\,.922\,/\,.200 & 22.40\,/\,.872\,/\,\nometric\\
& 3DGEER (+ footprint floor)
& 28.73\,/\,.949\,/\,.112 & \snd{28.13}\,/\,\snd{.921}\,/\,\snd{.213} & \snd{27.42}\,/\,\snd{.897}\,/\,\snd{.271} & \snd{28.09}\,/\,\snd{.922}\,/\,\snd{.199} & \snd{22.82}\,/\,\snd{.879}\,/\,\nometric\\
& ConeGaussian (ours)
& \snd{28.80}\,/\,\snd{.950}\,/\,\sota{.111} & \sota{28.32}\,/\,\sota{.924}\,/\,\sota{.204} & \sota{27.52}\,/\,\sota{.900}\,/\,\sota{.264} & \sota{28.21}\,/\,\sota{.925}\,/\,\sota{.193} & \sota{22.88}\,/\,\sota{.879}\,/\,\nometric\\
\cmidrule(lr){1-7}
\multirow{4}{*}{\shortstack[l]{\emph{Zip-NeRF}\\[1pt]\scriptsize fisheye}}
& 3DGEER (baseline)
& 25.87\,/\,.899\,/\,.150 & 24.92\,/\,.845\,/\,\snd{.278} & 24.13\,/\,.787\,/\,.409 & 24.97\,/\,.844\,/\,.279 & 21.72\,/\,.780\,/\,\nometric\\
& 3DGEER (+ Mip-Splatting floor)
& 25.88\,/\,.899\,/\,.150 & 24.92\,/\,.845\,/\,.279 & 24.13\,/\,.787\,/\,.409 & 24.98\,/\,.844\,/\,.279 & \snd{21.73}\,/\,.779\,/\,\nometric\\
& 3DGEER (+ footprint floor)
& \snd{25.91}\,/\,\snd{.900}\,/\,\snd{.150} & \snd{24.94}\,/\,\snd{.845}\,/\,.279 & \snd{24.14}\,/\,\snd{.787}\,/\,\snd{.409} & \snd{25.00}\,/\,\snd{.844}\,/\,\snd{.279} & 21.72\,/\,\snd{.780}\,/\,\nometric\\
& ConeGaussian (ours)
& \sota{25.96}\,/\,\sota{.901}\,/\,\sota{.148} & \sota{25.02}\,/\,\sota{.851}\,/\,\sota{.270} & \sota{24.25}\,/\,\sota{.795}\,/\,\sota{.398} & \sota{25.08}\,/\,\sota{.849}\,/\,\sota{.272} & \sota{21.74}\,/\,\sota{.781}\,/\,\nometric\\
\cmidrule(lr){1-7}
\multirow{4}{*}{\shortstack[l]{\emph{Mip-NeRF 360}\\[1pt]\scriptsize pinhole}}
& 3DGEER (baseline)
& 29.73\,/\,.898\,/\,.128 & 26.92\,/\,.790\,/\,.255 & 25.62\,/\,.720\,/\,.379 & 27.42\,/\,.803\,/\,.254 & 25.77\,/\,.718\,/\,\nometric\\
& 3DGEER (+ Mip-Splatting floor)
& 29.93\,/\,.902\,/\,.125 & 27.03\,/\,.794\,/\,.252 & 25.74\,/\,.727\,/\,.372 & 27.57\,/\,.808\,/\,.250 & \snd{25.94}\,/\,\snd{.726}\,/\,\nometric\\
& 3DGEER (+ footprint floor)
& \snd{30.05}\,/\,\snd{.906}\,/\,\snd{.118} & \snd{27.11}\,/\,\snd{.798}\,/\,\snd{.243} & \snd{25.77}\,/\,\snd{.728}\,/\,\snd{.365} & \snd{27.64}\,/\,\snd{.811}\,/\,\snd{.242} & \sota{25.96}\,/\,\snd{.726}\,/\,\nometric\\
& ConeGaussian (ours)
& \sota{30.15}\,/\,\sota{.907}\,/\,\sota{.117} & \sota{27.18}\,/\,\sota{.802}\,/\,\sota{.239} & \sota{25.79}\,/\,\sota{.731}\,/\,\sota{.359} & \sota{27.71}\,/\,\sota{.813}\,/\,\sota{.238} & \sota{25.96}\,/\,\sota{.729}\,/\,\nometric\\
\bottomrule
\end{tabular}
}
\caption{\textbf{Zoom-in}.
Models are trained at the lowest resolution and rendered at
$\times1$--$\times4$ magnification. Corner PSNR/SSIM are computed over the normalized radial
band r > 0.9.}
\vspace{-2em}
\label{tab:zoomin}
\end{table*}

\paragraph{MTMT}
\begin{figure}[htpb]
\vspace{-1em}
  \centering
  \includegraphics[
    width=0.48\textwidth
  ]{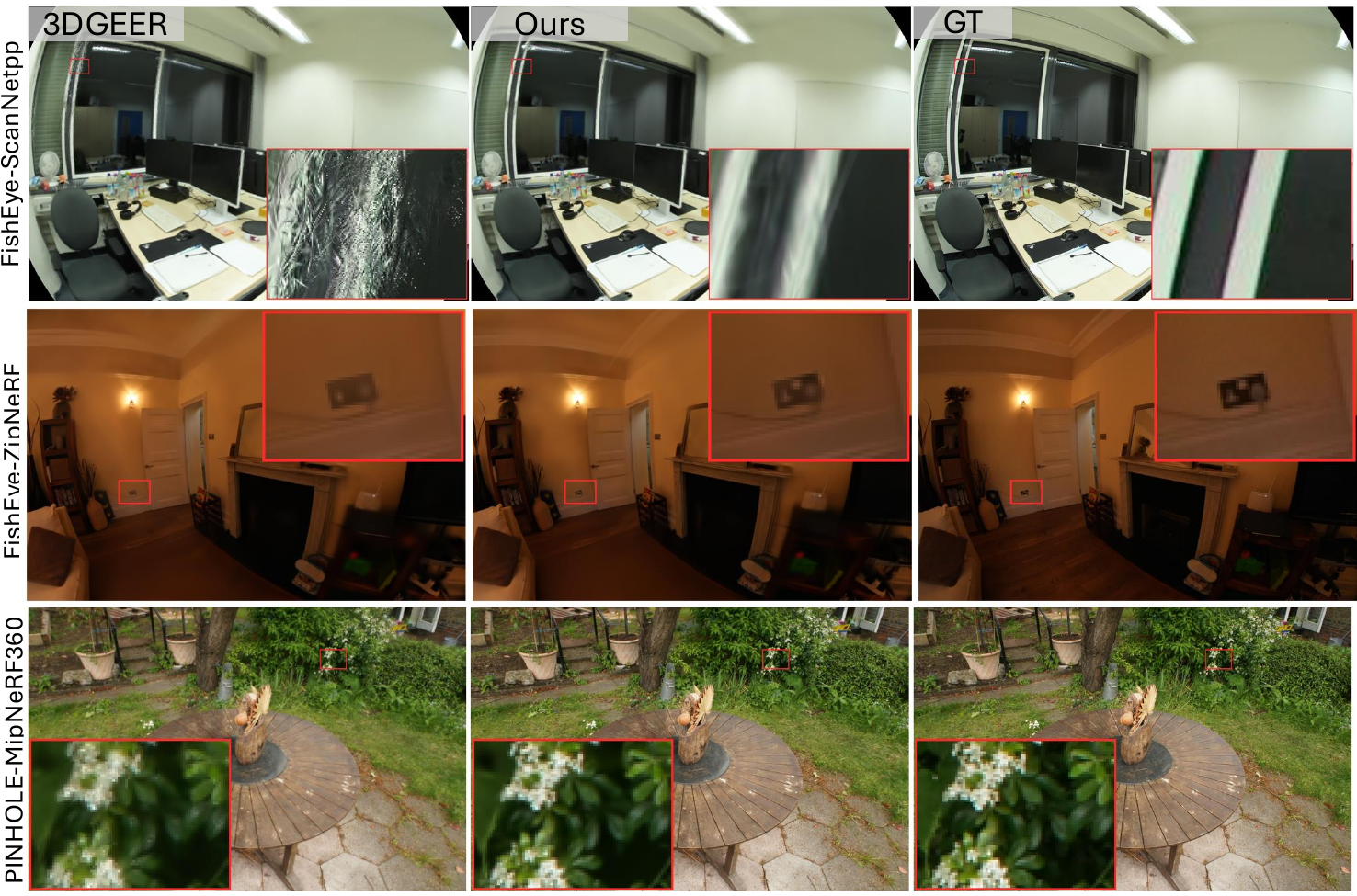}
  \vspace{-1.5em}
  \caption{
    {\textbf{Qualitative Results.} Comparisons across ScanNet++, Zip-NeRF, and Mip-NeRF 360, covering fisheye and pinhole cameras.}
  }
  \vspace{-1em}
  \label{fig:quali}
\end{figure}
Table~\ref{tab:ngauss} shows that \ConeGaussian{} {improves the scale-averaged metrics of both}
ray-based Gaussian renderers across fisheye and pinhole datasets, confirming
portability across ray--Gaussian formulations and camera models. The largest
gains occur at lower resolutions, where minification is strongest.
{Fig.~\ref{fig:quali} illustrates two types of improvement.
In the ScanNet++ window-frame crop, our result suppresses the baseline's
spurious streaks around an otherwise smooth structure. In the Zip-NeRF
wall-socket and Mip-NeRF 360 foliage crops, the socket outline and leaf
boundaries are more distinct and closer to the ground truth.} The
pinhole-specific screen-space methods use different rendering backbones and
are therefore included only as references; \ConeGaussian{} reaches comparable
multi-scale quality while retaining support for distorted fisheye cameras.

\begin{figure}[htpb]
  \centering
  \includegraphics[
    width=0.48\textwidth
  ]{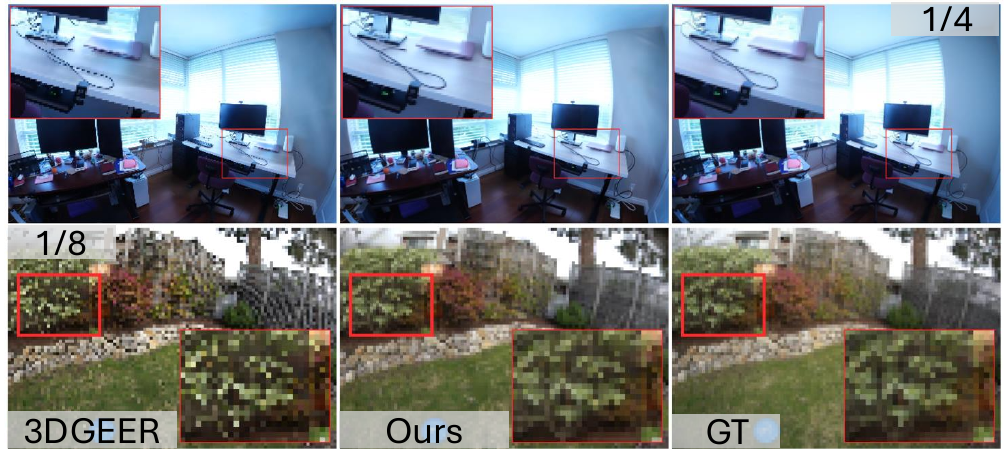}
  \vspace{-1.5em}
  \caption{
    {\textbf{Zoom-out Qualitative Results.} Highlighted cable and foliage crops compare the baseline, ours, and ground truth.}
  }
  \vspace{-1em}
  \label{fig:zoomout}
\end{figure}

\paragraph{Zoom-out}
The isotropic and anisotropic variants both disable the training-frequency floor, isolating the contribution of render-time footprint filtering under zoom-out. We additionally reimplement only the VKRayGS $z/f$ footprint-sizing component within the same framework, rather than its complete renderer.{Table~\ref{tab:zoomout} shows that the anisotropic footprint
achieves the highest average PSNR and lowest average LPIPS on all three
datasets. The $z/f$ approximation degrades under strong minification on ScanNet++
and Mip-NeRF 360. Fig.~\ref{fig:zoomout} compares the baseline with our filtered rendering.
In the indoor crop, the baseline breaks the thin cable into separated dark
samples, whereas ours renders a more continuous curve. In the outdoor crop, ours reduces isolated bright samples in the
foliage and better matches the lower-contrast appearance of the ground truth.}

\paragraph{Zoom-in}

\begin{table*}[htbp]
\centering\small
\setlength{\tabcolsep}{3pt}
\begin{tabular}{ll ccccc}
\toprule
Setting & Method & $1$ & $\tfrac12$ & $\tfrac14$ & $\tfrac18$ & Avg.\\
\midrule
\multirow{3}{*}{\shortstack[l]{FH\,$\to$\,PH\\[1pt]\scriptsize full img}}
& 3DGEER (baseline)
& \sota{27.42}\,/\,\sota{.935}\,/\,.257 & \snd{29.99}\,/\,\snd{.940}\,/\,.203
& \snd{29.66}\,/\,\snd{.941}\,/\,\snd{.156} & \snd{28.34}\,/\,\snd{.930}\,/\,\snd{.136}
& \snd{28.85}\,/\,\snd{.936}\,/\,\snd{.188}\\
& VKRayGS-insp.\ $z/f$ dilation
& 26.99\,/\,.933\,/\,\sota{.253} & 28.50\,/\,.937\,/\,\snd{.202}
& 25.73\,/\,.919\,/\,.172 & 23.02\,/\,.871\,/\,.172
& 26.06\,/\,.915\,/\,.200\\
& ConeGaussian (ours)
& \snd{27.38}\,/\,\snd{.934}\,/\,\snd{.255} & \sota{30.14}\,/\,\sota{.942}\,/\,\sota{.199}
& \sota{30.28}\,/\,\sota{.949}\,/\,\sota{.135} & \sota{29.56}\,/\,\sota{.951}\,/\,\sota{.083}
& \sota{29.34}\,/\,\sota{.944}\,/\,\sota{.168}\\
\cmidrule(lr){1-7}
\multirow{2}{*}{\shortstack[l]{PH\,$\to$\,FH\\[1pt]\scriptsize full img}}
& 3DGEER (baseline)
& \sota{27.01}\,/\,.919\,/\,.234 & \sota{27.47}\,/\,\sota{.931}\,/\,.185
& 27.53\,/\,.936\,/\,.139 & 26.66\,/\,.923\,/\,.120
& 27.17\,/\,.927\,/\,.169\\
& ConeGaussian (ours)
& \sota{27.01}\,/\,\sota{.920}\,/\,\sota{.230} & \sota{27.47}\,/\,\sota{.931}\,/\,\sota{.181}
& \sota{27.69}\,/\,\sota{.942}\,/\,\sota{.124} & \sota{27.02}\,/\,\sota{.939}\,/\,\sota{.077}
& \sota{27.30}\,/\,\sota{.933}\,/\,\sota{.153}\\
\cmidrule(lr){2-7}
\multirow{2}{*}{\shortstack[l]{PH\,$\to$\,FH\\[1pt]\scriptsize peripheral img}}
& 3DGEER (baseline)
& 23.82\,/\,.892\,/\,\nometric & 24.11\,/\,.897\,/\,\nometric
& 24.32\,/\,.901\,/\,\nometric & 24.12\,/\,.890\,/\,\nometric
& 24.09\,/\,.895\,/\,\nometric\\
& ConeGaussian (ours)
& \sota{23.83}\,/\,\sota{.893}\,/\,\nometric & \sota{24.15}\,/\,\sota{.898}\,/\,\nometric
& \sota{24.44}\,/\,\sota{.907}\,/\,\nometric & \sota{24.36}\,/\,\sota{.904}\,/\,\nometric
& \sota{24.20}\,/\,\sota{.900}\,/\,\nometric\\
\bottomrule
\end{tabular}
\caption{\textbf{Cross-camera rendering} on ScanNet++
(FH\,=\,fisheye, PH\,=\,pinhole).
FH$\to$PH models are trained on native fisheye views and rendered through the matched pinhole cameras without retraining; PH$\to$FH evaluates the reverse transfer at the same four scales. Peripheral PSNR/SSIM are computed over the normalized radial band
$r>0.7$.}
\label{tab:crosscam}
\end{table*}

\begin{table*}[htbp]
\centering\footnotesize
\setlength{\tabcolsep}{3pt}
\begin{tabular}{l cccccc}
\toprule
Method & $\times4$ & $\times2$ & $\times1$ & $\tfrac12$ & $\tfrac14$ & $\tfrac18$\\
\midrule
ConeGaussian (ours)
& \sota{27.57}\,/\,\sota{.900}\,/\,\sota{.264}
& \sota{28.37}\,/\,\sota{.924}\,/\,\sota{.204}
& \sota{28.86}\,/\,\sota{.950}\,/\,\snd{.111}
& \sota{28.94}\,/\,\sota{.959}\,/\,\sota{.052}
& \sota{29.22}\,/\,\sota{.960}\,/\,\sota{.041}
& \snd{27.18}\,/\,\snd{.949}\,/\,\sota{.044}\\
\;additive combination
& 27.37\,/\,\sota{.900}\,/\,\sota{.264}
& 28.14\,/\,\snd{.923}\,/\,\snd{.205}
& \snd{28.80}\,/\,\sota{.950}\,/\,\sota{.110}
& \snd{28.64}\,/\,\snd{.958}\,/\,\snd{.053}
& \snd{29.20}\,/\,\snd{.959}\,/\,\sota{.041}
& \sota{27.24}\,/\,\sota{.950}\,/\,\sota{.044}\\
\;w/o 3D floor
& \snd{27.47}\,/\,\snd{.898}\,/\,.272
& \snd{28.24}\,/\,\snd{.923}\,/\,.207
& 28.79\,/\,\snd{.949}\,/\,.113
& 28.14\,/\,.956\,/\,.056
& 29.10\,/\,.958\,/\,\snd{.044}
& 27.02\,/\,.947\,/\,\snd{.046}\\
\;w/o 2D filter
& 27.46\,/\,.897\,/\,\snd{.271}
& 28.18\,/\,.921\,/\,.213
& 28.78\,/\,\snd{.949}\,/\,.112
& 28.47\,/\,.951\,/\,.069
& 27.09\,/\,.930\,/\,.092
& 24.30\,/\,.901\,/\,.096\\
\;w/o both (3DGEER)
& 27.37\,/\,.895\,/\,.279
& 27.78\,/\,.918\,/\,.221
& 27.34\,/\,.944\,/\,.122
& 28.36\,/\,.950\,/\,.074
& 27.04\,/\,.929\,/\,.095
& 24.31\,/\,.901\,/\,.096\\
\bottomrule
\end{tabular}
\caption{\textbf{Joint module ablation} on ScanNet++ fisheye.
Models are trained at $\tfrac14$ native resolution and evaluated from
$\times4$ to $\tfrac18$ relative to the training scale.}
\label{tab:ablation22}
\vspace{-1em}
\end{table*}

\begin{figure}[htbp]
  \centering
  \includegraphics[
    width=0.48\textwidth
  ]{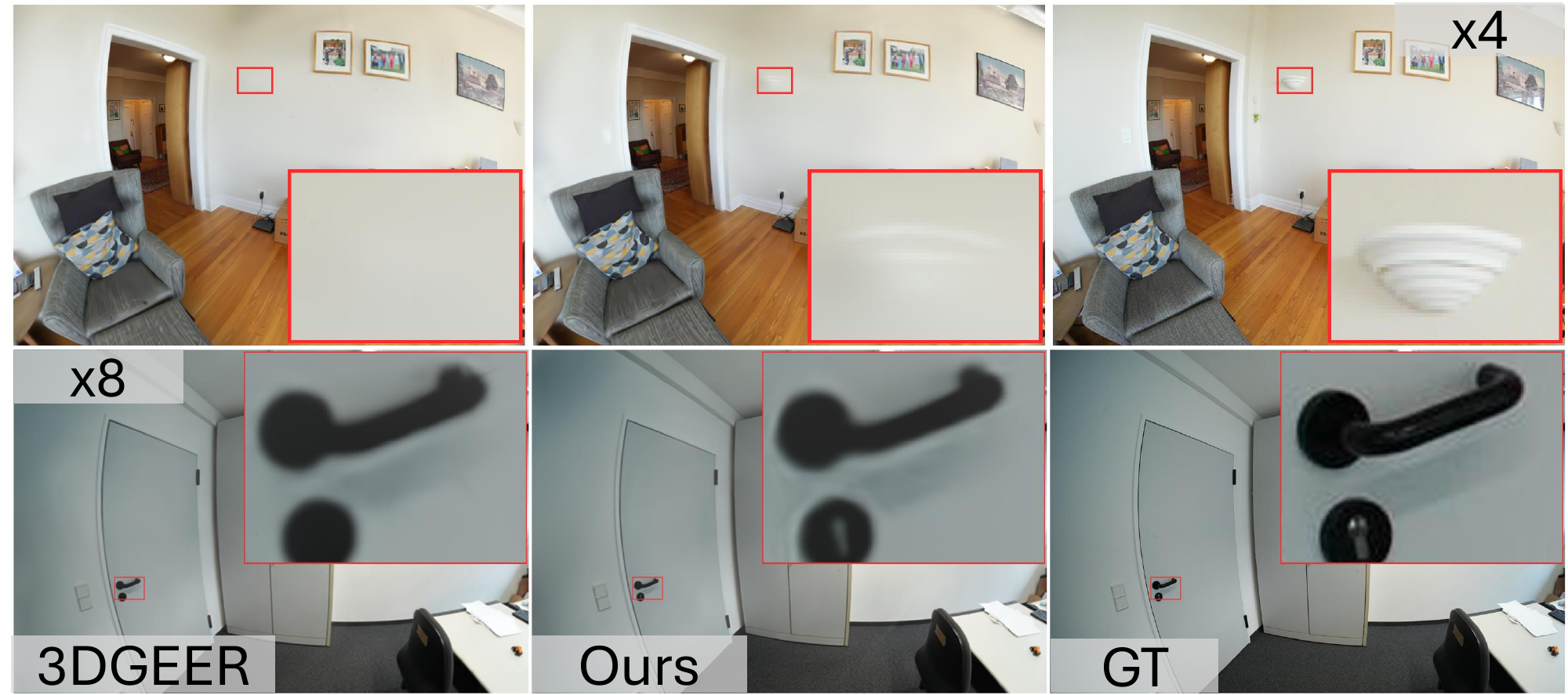}
  \vspace{-1.5em}
  \caption{
    {\textbf{Zoom-in Qualitative Results.} Highlighted door-handle and wall-light crops compare the baseline, ours, and ground truth.}
  }
  \vspace{-1em}
  \label{fig:zoomin}
\end{figure}

{Table~\ref{tab:zoomin} compares the paraxial $z/f$ floor,
our ray-derived footprint floor, and its marginal composition with the
render-time filter within the same ray-based renderer. The complete model
achieves the best average PSNR, SSIM, and LPIPS across the evaluated datasets.
Fig.~\ref{fig:zoomin} shows a more distinct keyhole below the door handle
and a faint wall-light structure that is nearly absent in the baseline.
These improvements preserve recognizable structure under magnification. More qualitative visualization is shown in the supplementary material.}

\subsection{Cross-camera rendering}
\label{sec:crosscam}

Cross-camera rendering changes the pixel ray bundles while
keeping the learned scene fixed. Table~\ref{tab:crosscam} evaluates whether
recomputing the footprint from the target camera preserves rendering quality
without retraining. In both transfer directions, the gains are most evident
under minification, consistent with the greater need for footprint integration
as each pixel covers a wider bundle of rays. The VKRayGS-inspired $z/f$
sizing instead deteriorates in this regime for FH$\to$PH, suggesting that
camera transfer requires appropriate footprint sizing, not simply more
smoothing. For PH$\to$FH, improvements extend to the distorted peripheral
band at every tested scale. This pattern supports adapting the filter to the target
camera's local ray geometry, with the clearest benefit when sampling becomes
coarse. The results demonstrate transfer between the tested pinhole and
fisheye cameras without modifying the learned scene representation.

\subsection{Ablation study}
\label{sec:ablation}
\label{sec:freqfilter}

We ablate the two band-limits and the footprint construction in
Table~\ref{tab:ablation22}, with the marginal-vs-additive composition analyzed
further in the Appendix; we summarize the findings
here. The
render-time footprint filter and the training-frequency floor are
complementary and both necessary: the filter carries the minification gains
(removing it collapses zoom-out by up to $2.9$\,dB). {
The render-time filter can also affect magnification results through its
effect on the learned representation during training, so these roles do not
imply that the two modules act exclusively in separate scale regimes.}
Among floor \emph{sizings}, the footprint floor matches Mip-Splatting's
$z/f$ on average and is more stable in the distorted periphery; decoupling
view selection from footprint sizing (a ray-based footprint at a single
$z/f$-selected view) collapses the training scale, so the floor must be the
\emph{minimum} of the \emph{true} footprint over training views. The anisotropic footprint is most beneficial where footprints shear:
it yields $5$--$10\times$ lower closed-form error and improves performance.

\section{Conclusion}

We presented \ConeGaussian{}, an approach for training and rendering 3DGS with ray-based central camera models while accurately modeling the pixel footprint in the scene space. Our main result is a closed-form anisotropic pixel footprint filter properly sized at the
max-response depth, generated by the
same camera model as the rays themselves. Importantly, because the footprint is built from the camera's own inverse ray mapping, it ports across arbitrary central camera models. On the other hand, the footprint filter acts only across the ray, leaving the along-ray response untouched, allowing seamless inclusion of any ray-Gaussian backbone. We provide theoretical derivations and error-analysis of the proposed footprint filter. At render time, our scene-space footprint representation allows direct composition of the footprint filter with the training-frequency floor through the proposed marginal composition rule, removing excess blurring. We validate our method in fish-eye and pin-hole camera models in multiple resolutions with two different ray-Gaussian backbones. Our method consistently and significantly improve each ray-Gaussian backbone in several experimental settings. Nevertheless, a limitation of our formulation is that it handles only calibrated central camera models whose with single optical center. Other interesting cases, such as catadioptric cameras and rolling-shutter capture, are left to future work.

\newpage
\clearpage
\bibliography{aaai2027}

\end{document}